# Extended Lifted Inference with Joint Formulas

**Udi Apsel** and **Ronen I. Brafman**
Computer Science Dept.
Ben-Gurion University of The Negev
Beer-Sheva, Israel 84105
apsel,brafman@cs.bgu.ac.il

## Abstract

The First-Order Variable Elimination (FOVE) algorithm allows exact inference to be applied directly to probabilistic relational models, and has proven to be vastly superior to the application of standard inference methods on a grounded propositional model. Still, FOVE operators can be applied under restricted conditions, often forcing one to resort to propositional inference. This paper aims to extend the applicability of FOVE by providing two new model conversion operators: the first and the primary is *joint formula conversion* and the second is *just-different counting conversion*. These new operations allow efficient inference methods to be applied directly on relational models, where no existing efficient method could be applied hitherto. In addition, aided by these capabilities, we show how to adapt FOVE to provide exact solutions to Maximum Expected Utility (MEU) queries over relational models for decision under uncertainty. Experimental evaluations show our algorithms to provide significant speedup over the alternatives.

## 1 Introduction

Probabilistic graphical models have been widely used over the last two decades in real-world and research applications. One of their sought after features is the ability to compactly represent a set of interdependencies among random variables, providing a platform for efficient inference methods for both exact [1] and approximate [16] inference.

Probabilistic Relational Models (PRM) extend the propositional models by introducing the concept of domain entities, along with a richer language which depicts the properties of each entity and the various interactions which they exhibit. Naturally, it is desirable, and often much more efficient, to apply inference directly to the relational model, thus avoiding an explicit extraction of the propositional model. The act of exploiting the high level structure in relational models is called *lifted inference*. This task can be carried out by a family of exact lifted inference algorithms, which are based on the idea of *First-Order Variable Elimination* (FOVE) [12, 3, 9].

An important task which is closely related to probabilistic inference, is *decision making under uncertainty*. The tight connection between the two tasks is exemplified in the *influence diagram* model [7], a popular model for decision making. Influence diagrams extend probabilistic models by adding decision and utility components to probabilistic graphical models. The quality of a decision, a set of assignments to decision variables in the influence diagram, is measured by its Expected Utility (EU). Under this principle, the best decision is achieved by maximizing the expected utility, a task that has been studied for both exact resolution [5] and approximation [11]. In the relational models realm, the study of decision making in influence diagrams has focused mainly on first-order MDP [14].

The goal of this paper is to extend the applicability of FOVE in two directions. First, we enrich the set of operators used by FOVE, by (a) introducing a novel model conversion method called *joint formula conversion*, and (b) generalizing the known *counting conversion* [9] operator to support the conversion of *just-different atoms* [4]. Joint formula conversion is a procedure which couples together pairs of atomic formulas, by replacing all their occurrences in the model with a new formula, whose range is a Cartesian product of the original pair. As we explain and demonstrate empirically, the conversion allows a subsequent use of efficient inference operators: *counting conversion* [9] and *inversion* [3], where previously one would resort to grounding. Additionally, the combination of (a) and (b) allows further lifting in cases that were previously considered hard for lifted inference.

Second, we present a solution to decision making in first-order influence diagrams [7] based on the FOVE algorithm, the first lifted solution to the best of our knowledge. Our method applies a variation of C-FOVE [9] that computes maximum expected utility (MEU) [5]. We show that variations of counting conversion and inversion can lift the MEU computation, much like in the belief assessment and MPE tasks. Similarly to other FOVE variations, experimental evaluations show our lifted method to be substantially more efficient than the propositional alternative.

We note that recent works [6, 8, 15] demonstrate the advantage of exploiting the logical structure of first-order formulas (e.g. MLN features [13], preference rules [2]) for the benefit of efficient lifted inference. FOVE, on the other hand, operates under no assumption on the logical structure of the first-order formulas which compose the relational model. A comprehensive comparison study between these different approaches has yet to be conducted.

## 2 Model Representation

Based on Markov Logic Decision Network (MLDN) [10] and the work of Milch et al. [9], we present a first-order model which depicts two types of variables: *random variables* and *decision variables*, and two types of factors – *probability factors* and *utility factors*.

### 2.1 Atoms, Constraints and Parfactors

Each variable induced by the model corresponds to a *ground atom* of the form $p(c_1, \ldots, c_n)$, where $p$ is a predicate of finite range, $range(p)$, and $c_1, \ldots, c_n$ are constant symbols. An atomic formula $p(t_1, \ldots, t_n)$ where $t_i$ is a constant or a *logical variable*, is called an $atom$. Each logical variable $X$ is bound by a domain $dom(X)$ with cardinality $|dom(X)|$, or $|X|$. $LV(\alpha)$ is the set of logical variables referred by $\alpha$, where $\alpha$ is a formula or a set of formulas. Under a set of assignments $v$, the notation $\alpha(v)$ is used to depict the values assigned to $\alpha$.

A *factor* $f$ is a pair $(A, \eta)$, consisting of a set of ground formulas and a potential function $\eta : \prod_{\alpha \in A} range(\alpha) \to \mathbb{R}$. Under a set of assignments $v$, the weight of factor $f$ is $w_f(v) = \eta(\alpha_1(v), \ldots, \alpha_m(v))$, where $A = \{\alpha_1, \ldots, \alpha_m\}$. A *substitution* $\theta$ over a set of logical variables $L$ maps each variable in $L$ to a constant symbol or a logical variable. $\alpha\theta$ depicts the result of applying a substitution $\theta$ on $\alpha$.

A constraint $C$ is a pair $(F, L)$, where $F$ is an equational formula on logical variables set $L$. $gr(L : C)$ is a set of substitutions on $L$ under constraint $C$, where all logical variables of $L$ are substituted with a constant. Similarly to previous work [9], we require the constraints to be in some normal form, where for each logical variable $X$, $|X : C|$ has a fixed value regardless of the binding of other logical variables in $C$. We use $var(\alpha)$ to depict the set of variables specified by $\alpha$ under the set of substitutions $gr(L : C)$, and in-order to distinguish between the two types of variables in $var(\alpha)$, $rv(\alpha)$ is used to depict the set of random variables in $\alpha$, and $dv(\alpha)$ depicts the set of decision variables.

A *parfactor* $g$ is a tuple $(L, C, A, \eta)$, comprised of a set of logical variables, a constraint on $L$, a set of formulas and a potential, respectively. Applying a substitution $\theta$ over parfactor $g$ results in $g' = (L', C', A\theta, \eta)$, where $L'$ and $C'$ are obtained by applying substitution on its logical variables, and dropping those mapped to constants. A *ground substitution* of a parfactor is a factor which was generated by a substitution over all the logical variables. The model contains two types of parfactors, probability and utility, which depict a set of probability and utility factors upon grounding. As a convention, $\phi$ depicts a potential of a probability parfactor, and $\mu$ depicts a potential of a utility parfactor.

The weight of parfactor $g$, depicted by $w_g(v)$, is determined according to its type. The weight of a probability parfactor is $w_g(v) = \prod_{f \in gr(g)} w_f(v)$, and the weight of a utility parfactor is $w_g(v) = \sum_{f \in gr(g)} w_f(v)$. For convenience and clarity, we use the abbreviation $\eta(\alpha_1(L_1), \ldots, \alpha_i(L_i), C)$ to represent a parfactor constrained by $C$, which contains a set of formulas $\alpha_1, \ldots, \alpha_i$ with their respective variable scopes $L_1, \ldots, L_i$. For instance, the notation $\phi_1(s(X), t(Y, X), \{X \neq Y\})$ represents a probability parfactor whose properties are $L = \{X, Y\}, C = (\{X \neq Y\}, \{X, Y\}), A = \{s, t\}$, and $\eta = \phi_1$. An alternative notation for constraint $C$ is $C_{X \neq Y}$.

### 2.2 Counting Formulas and Histograms

*Counting formulas* express a numerical distribution of values on a portion of a formula's groundings, by counting the number of groundings that are assigned each possible value. Instead of covering each possible assignment, the counting formulas are oblivious to the specific permutations which conform to the count formation. The notation of counting formulas is $\#_{X:C}[\alpha]$ where $\alpha$ is the counted atom, $X$ is the counted logical variable, and $C$ is the parfactor's constraint over the counted population. For example, formula $\#_{Y:\{X \neq Y\}}[friends(X, Y)]$ counts the $Y$ population of any given $X$ in atom $friends(X, Y)$, under constraint $X \neq Y$. The range of a counting formula $\gamma = \#_{X:C}[\alpha]$, depicted by $range(\gamma)$, is a set of all possible *histograms*. A histogram is a set of non-negative integer counters, each corresponding to a specific assignment in $range(\alpha)$, where the sum of all counters is $|X : C|$.

## 3 Joint Formula Conversion

### 3.1 Definition

A *joint formula* is a composite of two formulas (atoms or counting formulas), whose range of assignments is a Cartesian product of the range of its components. For example, $j(X, Y) = \langle a(X, Y), b(Y, X) \rangle$ depicts a joint formula of atoms $a(X, Y)$ and $b(Y, X)$, over logical variables $X$ and $Y$. If $a$ and $b$ are boolean atoms with $range(a) = range(b) = \{0,1\}$, then $range(j) = \{0,1\} \times \{0,1\}$. The joined formulas must be of the same type. Namely, both must be decision formulas or random variable formulas.

**Table 1:** *Joint formula $j(X,Y) = \langle a(X,Y), b(Y,X) \rangle$*

| $a(X,Y)$ | $b(Y,X)$ | $\phi$ |
|---|---|---|
| 0 | 0 | 0.2 |
| 0 | 1 | 0.3 |
| 1 | 0 | 0.7 |
| 1 | 1 | 0.7 |

| $j(X,Y)$ | $\phi''$ |
|---|---|
| $\langle 0,0 \rangle$ | 0.2 |
| $\langle 0,1 \rangle$ | 0.3 |
| $\langle 1,0 \rangle$ | 0.7 |
| $\langle 1,1 \rangle$ | 0.7 |

| $p(Z)$ | $a(X,Y)$ | $\mu$ |
|---|---|---|
| 0 | 0 | 7 |
| 0 | 1 | 3 |
| 1 | 0 | 2 |
| 1 | 1 | 5 |

| $p(Z)$ | $j(X,Y)$ | $\mu'$ |
|---|---|---|
| 0 | $\langle 0,0 \rangle$ | 7 |
| 0 | $\langle 0,1 \rangle$ | 7 |
| 0 | $\langle 1,0 \rangle$ | 3 |
| 0 | $\langle 1,1 \rangle$ | 3 |
| 1 | $\langle 0,0 \rangle$ | 2 |
| 1 | $\langle 0,1 \rangle$ | 2 |
| 1 | $\langle 1,0 \rangle$ | 5 |
| 1 | $\langle 1,1 \rangle$ | 5 |

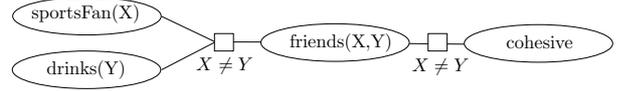

**Figure 1:** *A Markov logic network for the group cohesiveness problem: X and Y represent individuals of the same group, friendships are defined on pairs of individuals s.t. $X \neq Y$.*

*Joint formula conversion* is the replacement of all instances of a joint formula's components with the joint formula itself. Similarly to *shattering* [3], it can be applied at the beginning or during the inference task. The conversion conserves the assignment space of the original model, such that each assignment to ground atoms in the original model is mapped to a single respective assignment in the converted model, and vise versa. Both assignments, in the original model and in the converted model, yield the same results in all parfactors. For example, in a two parfactor model $\phi(a(X,Y), b(Y,X))$ and $\mu(p(Z), a(X,Y))$, a joint formula conversion for $j(X,Y) = \langle a(X,Y), b(Y,X) \rangle$ converts the model to $\phi'(j(X,Y), j(X,Y))$ and $\mu'(p(Z), j(X,Y))$, such that under each assignment $\langle v_a, v_b \rangle$ to a ground of $j$, the converted potentials yield the same values as their original counterparts under assignments $v_a$ and $v_b$ to grounds of $a$ and $b$, respectively. Parfactor $\phi'$ is compressed further to $\phi''(j(X,Y))$, since it contains two identical instances of $j$. The example is illustrated in Table 1.

### 3.2 Motivation and Example Applications

In a sense, joint formula conversion is counter-intuitive. Most lifting operators aim to reduce the variable assignment space, or to restructure the model without introducing unnecessary dependencies between variables. Joint formula conversion does the opposite – it deliberately introduces dependencies between formulas. However, this modification in structure may allow the inference task to benefit from lifting operators that would not be used otherwise. More specifically, joint formula conversion is highly efficient in cases where lifting operators are well defined on the joint formula, but not applicable on the separate components of the joint formula.

Let us demonstrate this with the task of summing-out all the random variables from a given parfactor, $\phi(a(X,Y), b(Y,X), c(X,Z), d(Z))$. Since both counting conversion and inversion are inapplicable in this case, some grounding operation must be applied. However, this overhead can be avoided by applying a joint formula conversion, for which $j(X,Y) = \langle a(X,Y), b(Y,X) \rangle$. The conversion yields parfactor $\phi'(j(X,Y), c(X,Z), d(Z))$, which in turn can be resolved by a sequence of lifting operators: (a) Applying counting conversion over $j(X,Y)$ w.r.t. $Y$. (b) Eliminating $c(X,Z)$ by inversion. (c) Applying counting conversion over $d(Z)$ w.r.t. $Z$. (d) Eliminating $\#_Y[j(X,Y)]$ by inversion. (e) Eliminating $\#_Z[d(Z)]$ by inversion. The amount of work that was invested in the joint formula conversion is therefore negligible compared with the overall computational benefit.

In Section 4, we introduce a variant of counting conversion which allows the conversion of *just-different atoms*. This newly introduced variation, combined with joint formulas, extends the scope of lifted inference in FOVE even further. For example, Figure 1 presents the group cohesiveness problem, where each member of a given group is examined according to two characteristics: affinity to sports and affinity to alcohol. The problem can be represented by two parfactors: $\phi_1(sportsFan(X), drinks(Y), friends(X,Y), C_{X \neq Y})$ – the chance of two individuals being friends, and $\phi_2(friends(X,Y), cohesive, C_{X \neq Y})$ – the chance of a group being cohesive.

In order to find out what are the chances of a group being cohesive, all variables but *cohesive* need to be summed-out from the model. We start by fusing $\phi_1$ and $\phi_2$ into $\phi(sportsFan(X), drinks(Y), friends(X,Y), cohesive, C_{X \neq Y})$, and eliminating $friends(X,Y)$ by inversion, resulting in $\phi'(sportsFan(X), drinks(Y), cohesive, C_{X \neq Y})$. Since counting conversion and inversion are both inapplicable in the model's current form, we apply a joint formula conversion with $j(X) = \langle sportsFan(X), drinks(X) \rangle$. The conversion results in $\phi''(j(X), j(Y), cohesive, C_{X \neq Y})$, and can be followed by a counting conversion of the $j$ instances, which are just-different atoms. Hence, the model is converted to $\phi'''(\#_X[j(X)], cohesive)$, and the inference task resumes without resorting to grounding.

### 3.3 Logical Variables Mapping

A *logical variables mapping* (or simply, mapping) between two formulas $\alpha$ and $\beta$, depicted by $M_{\alpha,\beta}$, is an isomorphism from the ordered set of logical variables of $\alpha$, $\vec{LV}(\alpha) = \langle \alpha_{[1]}, \ldots, \alpha_{[|LV(\alpha)|]} \rangle$, to the ordered set of logical variables of $\beta$, $\vec{LV}(\beta) = \langle \beta_{[1]}, \ldots, \beta_{[|LV(\beta)|]} \rangle$, where $\alpha_{[i]}$ and $\beta_{[j]}$ depict the i-th and j-th logical variables of $\alpha$ and $\beta$ under argument list ordering, respectively. Pairing of logical variables from $\alpha$ and $\beta$ is allowed only in cases where they have the same domain. For example, a possible mapping between $a(X,Y)$ and $b(W,Z)$ is $M_{a,b} = \{a_{[1]} \leftrightarrow b_{[2]}, a_{[2]} \leftrightarrow b_{[1]}\}$, provided that $dom(X) = dom(Z)$ and $dom(Y) = dom(W)$. We use $M_{\alpha \to \beta} : \vec{L}$ to depict a permutation of $\vec{L}$ according to the mapping from $\alpha$ variables to $\beta$ variables. In the given example, $M_{a \to b} : \langle Z, M \rangle = \langle M, Z \rangle$.

A *full mapping* between $\alpha$ and $\beta$ is a mapping over all the logical variables of both formulas, and a joint formula conversion is defined according to such a mapping. For example, in model $\phi(a(X,Y), b(Y,X))$, a joint formula conversion over mapping $M_{a,b} = \{a_{[1]} \leftrightarrow b_{[2]}, a_{[2]} \leftrightarrow b_{[1]}\}$ results in the joint formula $j(X,Y) = \langle a(X,Y), b(Y,X) \rangle$, and in a following conversion $\phi'(j(X,Y), j(X,Y))$, which can be simplified further to $\phi''(j(X,Y))$. On the other hand, a joint formula conversion of the same model over a different mapping, $M_{a,b} = \{a_{[1]} \leftrightarrow b_{[1]}, a_{[2]} \leftrightarrow b_{[2]}\}$, results in the conversion $\phi'(j(X,Y), j(Y,X))$, yielding no computational gain. Hence, joint formula conversions do not necessarily result in a more efficient inference, and their use should be considered only in cases where computational gain is guaranteed.

### 3.4 Usage and Computational Complexity

In the context of current C-FOVE implementations, where a greedy algorithm is used to determine which operator to apply next, joint formulas can simply be used when (a) all other lifting attempts fail, and (b) their placement allows subsequent counting conversions and inversions. However, given the proper heuristics, joint formula conversion can be applied at any phase of the inference task.

The computational complexity of joint formula conversion is bounded by $O(k \cdot r^{n+1})$, where $r$ is the maximum assignment range of any formula in the model, $k$ is the number of parfactors which consist of the joint formulas components, and $n$ is the maximum number of formulas in any of the subject parfactors.

### 3.5 Joint Shattering

Before applying a joint formula conversion, a *joint shattering* has to be carried-out. Joint shattering is identical to the already known shattering [3] process, only that the formulas which are about to be joined, $\alpha$ and $\beta$, are shattered w.r.t. their instances under the joint formula. Namely, the joint shattering splits the set of parfactors in the model, such that parfactors which contain $\alpha(\vec{L_\alpha})$ are treated as if they contained $\beta(\vec{L_\beta})$ as well, where $\vec{L_\beta} = M_{\alpha \to \beta} : \vec{L_\alpha}$. Similarly, parfactors which contain $\beta(\vec{L_\beta})$ are treated as if they contained $\alpha(\vec{L_\alpha})$. Let us demonstrate this with an example. Assume a model $\phi(a(X,Y), b(X,Z), C_{X \neq Z})$ which is about to be applied with a joint formula conversion over mapping $M_{a,b} = \{a_{[1]} \leftrightarrow b_{[1]}, a_{[2]} \leftrightarrow b_{[2]}\}$, where $dom(X) = dom(Y) = dom(Z) = \{x_1, x_2\}$. A joint formula $j(X,Y) = \langle a(X,Y), b(X,Y) \rangle$ cannot be placed in the model's current form, for two reasons. The first, is that there are no constraints which prevent an equality between $X$ and $Y$, hence $j(x_1, x_1)$ is a possible ground of $j$ in one of the converted parfactor's grounding. However, $j(x_1, x_1)$ implies that the set of random variables in the original model includes $b(x_1, x_1)$, which is untrue. A second reason is that the placement of $j$ would result in parfactor $\phi_j(j(X,Y), j(X,Z), C_{X \neq Z})$, where the two instances of $j$ entail two sets of ground variables which are neither disjoint nor equal.

A joint shattering of $\phi(a(X,Y), b(X,Z), C_{X \neq Z})$ treats the parfactor as if it contained both $a(X,Z)$ and $b(X,Y)$. In this case, the parfactor is split on substitution $Y/X$, where two parfactors are created: $\phi_1(a(X,X), b(X,Z), C_{X \neq Z})$, and $\phi_2(a(X,Y), b(X,Z)), C_{\{X \neq Y, X \neq Z\}})$. $a(X,X)$ here is practically a different formula than $a(X,Y)$ where $X \neq Y$, since the sets of grounds for both are disjoint. Placing the joint formula in the model's current form should yield parfactors $\phi_j^1(a(X,X), j(X,Z), C_{X \neq Z})$ and $\phi_j^2(j(X,Y), j(X,Z)), C_{\{X \neq Y, X \neq Z\}})$.

## 4 Just-Different Counting Conversion

The notion of *just-different atoms* was introduced by Braz et al. [4] for the purpose of *counting elimination*, but has yet to be exploited for the purpose of *counting conversion*. As mentioned earlier, the combination of counting conversion of just-different atoms with joint formulas, extends the scope of lifted inference and provides motivation to explore this variation of counting conversion. For the purpose of simplicity and clarity, we present a version which converts pairs of just-different atoms. Note that the procedure can be generalized to any number of just-different atom. The simple case of counting conversion, where a single formula is converted, is directly derived from this more general case.

Let parfactor $g_\eta$ contain two instances of formula $\alpha$: $\alpha_1 = \alpha(X, L)$ and $\alpha_2 = \alpha(Y, L)$, where $X$ and $Y$ are logical variables, $L$ is a set of logical variables, $X \notin L$ and $Y \notin L$. Let any ground substitution of $L$ produce a set of just-different atoms, namely: for each given substitution of $L$, choosing one substitution of $X$ restricts $Y$ in only one substitution, and vice versa. An example of such a par-factor is $\phi(\alpha(Z,X), \alpha(Z,Y), C_{X \neq Y})$. Finally, let both $X$ and $Y$ be owned exclusively by their $\alpha$ instances, such that no other formula in $g$ contains neither $X$ nor $Y$.

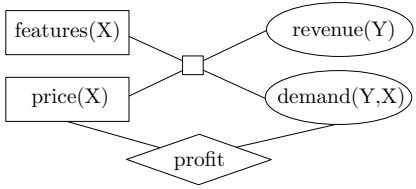

**Figure 2:** *An MLDN for product planning, depicting decision nodes (rectangles), uncertainty nodes (ellipses), and value nodes (diamonds).*

A counting conversion of formulas $\alpha_1$ and $\alpha_2$ over logical variables $X$ and $Y$ in parfactor $g_\eta$ is a conversion of $g_\eta = (L_\eta, C_\eta, A_\eta, \eta)$ to $g_{\eta'} = (L_{\eta'}, C_{\eta'}, A_{\eta'}, \eta')$, by replacing the two $\alpha$ instances with an arity-reduced counting formula $\#_{X:C_\eta}[\alpha]$, and defining a potential $\eta'$, such that in probability parfactors

$$\eta'(N, b_1, \ldots, b_k) = \prod_{a_1, a_2 \in range(\alpha)} \eta(a_1, a_2, b_1, \ldots, b_k)^{\#(N, a_1, a_2)} \quad (1)$$

and in utility parfactors

$$\eta'(N, b_1, \ldots, b_k) = \sum_{a_1, a_2 \in range(\alpha)} \eta(a_1, a_2, b_1, \ldots, b_k) \cdot \#(N, a_1, a_2) \quad (2)$$

Where

$$\#(N, a_1, a_2) = \begin{cases} \#(N, a_1) \cdot (\#(N, a_2) - 1) & a_1 = a_2 \\ \#(N, a_1) \cdot \#(N, a_2) & a_1 \neq a_2 \end{cases} \quad (3)$$

$a_1$ and $a_2$ are assignments to grounds of $\alpha_1$ and $\alpha_2$, $b_1, \ldots, b_k$ are assignments to grounds of all other formulas, and histogram $N = \{n_1, \ldots, n_r\}$ is a set of counters for each possible assignment of a ground of $\alpha$ under the conditions $r = |range(\alpha)|$ and $\sum_{i=1}^r n_i = |X:C|$. $\#(N, a)$ depicts the value of the entry which counts assignment $a$. The rest of $g_{\eta'}$ properties are obtained by $L_{\eta'} = L_\eta \setminus \{X, Y\}$, $A_{\eta'} = A_\eta \setminus \{\alpha_1, \alpha_2\}$, and $C_{\eta'} = C_\eta^{\downarrow}{L_{\eta'}}$ (the projection of the remaining logical variables). A number $comb(N)$ is then attributed to each histogram $N$

$$comb(N) = \frac{|X:C|!}{\prod_{a \in range(\alpha)} \#(N, a)!} \quad (4)$$

Where $comb(\langle \chi_1, \chi_2 \rangle) = comb(\chi_1) \cdot comb(\chi_2)$ in joint formulas, and $comb(\chi) = 1$ in atoms.

## 5 FOVE for MEU

To capture relational decision making settings, we use a model based on Markov Logic Decision Network (MLDN) [10]. The model includes probability and utility parfactors, and two types of formulas: random variable formulas and decision formulas.

### 5.1 MEU

Formally, $G = G_\phi \cup G_\mu$, where $G_\phi$ contains a set of probability parfactors, and $G_\mu$ contains a set of utility parfactors. The *expected utility* (EU) of model $G$ under assignment $v_d$ to (all) its decision variables is given by:

$$eu[G](v_d) = \frac{1}{Z} \sum_{rv(G)} \prod_{g \in G_\phi} w_g(v_d) \cdot \sum_{g \in G_\mu} w_g(v_d) \quad (5)$$

In our setting, we can ignore $Z$, which is the normalizing constant of the MLDN. The *maximum expected utility* (MEU) of model $G$ is given by

$$meu[G] = \left( \underset{v_d}{\operatorname{argmax}}\, eu[G](v_d),\, \max_{v_d} eu[G](v_d) \right) \quad (6)$$

To illustrate this model, consider Figure 2 which depicts a first-order decision problem, where a product planner has to decide on a line of products for the enterprise market. We seek a decision that maximizes the expect profit, i.e., one with maximum expected utility. The planner needs to determine each product's set of features and market price, and does so by examining the profile of each of the potential buyers – their yearly revenue and their demand for each of the expected products. In our model, the problem is represented by two parfactors: $\phi(features(X), price(X), revenue(Y), demand(Y,X))$ and $\mu(price(X), demand(Y,X))$. $\phi$ depicts probability weights of various interactions between variables, and $\mu$ depicts the utility portion (profit). The set of decision variables is represented by atoms $features(X)$ and $price(X)$, and the set of random variables is represented by $revenue(Y)$ and $demand(Y,X)$.

### 5.2 Challenges in Lifted MEU Computation

Lifted MEU introduces several challenges which do not exist in "normal" lifted inference. The first challenge stems from the presence of two types of formulas, decision and random variables, for which separate elimination procedures are defined. Notably, random variable atoms are eliminated by summing-out their effect on the network, whereas decision atoms are maximized-out from the network [5]. Consequently, the number $comb(N)$ which is typically attributed to each histogram $N$, serves no part in the elimination process of decision formulas. Additionally, decision formulas can be eliminated only from parfactors which contain no random variable formulas.

The second challenge arises from the two separate parts of the MEU expression, which depict the weights of two type of parfactors: probability and utility. The complex structure forces the inversion procedure to be more complicated than in belief assessment, but most importantly – it poses a significant restriction on the inversion of decision formulas: decision formulas can be eliminated by inversion only when contained in one type of parfactors. This restriction increases the importance of joint formula conversion, which allows counting conversion to be applied where normally such a use would not be allowed. Joint formulas are in no way a panacea for this inherent nature of the problem, however, without joint formulas many MEU computation tasks unnecessarily resort to propositionalization.

### 5.3 Framework

Given a model $G$, we begin by choosing which operator to apply. We have three lifting operators at our disposal: *inversion elimination*, *counting conversion* and *joint formula conversion*. We also have two grounding operators: *propositionalization* and *counting expansion*, which are carried-out identically to C-FOVE. After applying the operator of choice, we are left with a transformed model, $G'$, whose MEU solution entails the original model's MEU.

We continue to apply some operator of choice, repeatedly, until all remaining formulas are (a) decision formulas, and (b) ground formulas. Counting formulas with no active logical variables are considered to be ground formulas as well. Lastly, an exhaustive search is issued on the assignment space, in-order to find the maximizing assignments of the remaining ground formulas. A final backward phase, similar to the one used in lifted MPE [4], resolves the assignments of the eliminated decision formulas.

### 5.4 Inversion Elimination

Let $G_\alpha$ denote the set of parfactors which contain formula $\alpha$ in model $G$. Inversion elimination [3] of formula $\alpha$ can be applied to model $G$ under three conditions: (a) Model $G$ is shattered w.r.t. $\alpha$. (b) For each $g \in G_\alpha$, $\alpha$ contains all the logical variables of $g$. (c) The set of formulas in each $g \in G_\alpha$ contains only one instance of $\alpha$. Inversion eliminates $\alpha$ from the model and produces a residual model $G'$. During the elimination procedure, *product fusion* and *summation fusion* are repeatedly used, forming a parfactor with a single instance of $\alpha$ which contains all the logical variables of its container parfactor. Product fusion is defined in [3], and summation fusion is a similar procedure, with the distinction of summing potentials instead of applying multiplication. We now define formally, the procedure for eliminating random variable formulas, and the procedure for the elimination of decision formulas.

#### 5.4.1 Eliminating Random Variable Formulas

We assume formula $\alpha$ to reside in both probability and utility parfactors[1]. We start by merging all probability parfactors which contain $\alpha$ into $g_\phi = (L_\phi, C_\phi, A_\phi, \phi)$, using a product fusion. Let $g_\mu = (L_\mu, C_\mu, A_\mu, \mu)$ be some utility parfactor which contains $\alpha$, and let $g_\sigma = (L_\sigma, C_\sigma, A_\sigma, \sigma)$ be a product fusion of $g_\phi$ with $g_\mu$. Let $L_\phi^\alpha$ and $L_\sigma^\alpha$ denote the set of logical variables which are unique to $\alpha$ in parfactors $g_\phi$ and $g_\sigma$, respectively. A parfactor $g_{\phi'} = (L_{\phi'}, C_{\phi'}, A_{\phi'}, \phi')$ is obtained by calculating

$$\phi^{sum}(b_1, \ldots, b_k) = \sum_{a \in range(\alpha)} comb(a) \cdot \phi(a, b_1, \ldots, b_k) \quad (7)$$

Followed by

$$\phi'(b_1, \ldots, b_k) = \phi^{sum}(b_1, \ldots, b_k)^{|L_\phi^\alpha : C_\phi|} \quad (8)$$

Where $A_{\phi'} = A_\phi \setminus \{\alpha\}$, $L_{\phi'} = L_\phi \setminus L_\phi^\alpha$, and $C_{\phi'} = C_{\phi L_{\phi'}}^\downarrow$. As a convention, $b_1, \ldots, b_k$ depict $k$ assignments to all formulas in the subject parfactor, except formula $\alpha$. Next, for each of the $g_\mu$ parfactors, a respective $g_{\mu'} = (L_{\mu'}, C_{\mu'}, A_{\mu'}, \mu')$ is obtained by calculating

$$\sigma^{sum}(b_1, \ldots, b_n) = \sum_{a \in range(\alpha)} comb(a) \cdot \sigma(a, b_1, \ldots, b_n) \quad (9)$$

Followed by

$$\mu'(b_1, \ldots, b_n) = \frac{\sigma^{sum}(b_1, \ldots, b_n)}{\phi^{sum}(b_1, \ldots, b_k)} \cdot |L_\sigma^\alpha : C_\sigma| \quad (10)$$

Where $A_{\mu'} = A_\mu \setminus \{\alpha\}$, $L_{\mu'} = L_\mu \setminus L_\sigma^\alpha$, and $C_{\mu'} = C_{\mu L_{\mu'}}^\downarrow$. Note that $k \leq n$, since $A_\phi \subseteq A_\sigma$ as a result of $g_\sigma$ being a fusion of $g_\phi$ with $g_\mu$. Finally, a residual model $G'$ is obtained by replacing $g_\phi$ with $g_{\phi'}$, and replacing each of the $g_\mu$ parfactors with its respective $g_{\mu'}$.

Equations 8 and 10 instruct of exponentiation and multiplication in the combined domain sizes of the removed logical variables. In effect, these operations express the nature of inversion, where numerous variables are eliminated simultaneously. We demonstrate this with a two parfactor model $\phi(p(X), q(X, Y))$ and $\mu(r(Y), q(X, Y))$, for which we aim to eliminate random variable atom $q(X, Y)$. The elimination of $q(X, Y)$ is conducted in several steps. First, $\phi^{sum}$ is obtained by $\phi^{sum} = \sum_q \phi$. Since the elimination of $q$ removes logical variable $Y$ from parfactor $\phi(p(X), q(X, Y))$, $\phi'$ is obtained by $\phi' = (\phi^{sum})^{|Y|}$. Next, we fuse $\phi(p(X), q(X, Y))$ with $\mu(r(Y), q(X, Y))$, resulting in $\sigma(p(X), r(Y), q(X, Y))$. $\sigma^{sum}$ is then obtained by $\sigma^{sum} = \sum_q \phi \cdot \mu$. Here, a removal of $q$ from $\sigma(p(X), r(Y), q(X, Y))$ does not reduce the set of logical variables. Hence, $\mu'$ is obtained by $\mu' = \frac{\sigma^{sum}}{\phi^{sum}}$, without multiplication. A numerical example is given in Table 2.

#### 5.4.2 Eliminating Decision Formulas

Here, two additional precondition are required: (a) formula $\alpha$ is contained exclusively in utility parfactors or probability parfactors, but not in both. (b) All formulas which share a parfactor with $\alpha$ are decision formulas. Next, all parfactors which contain $\alpha$ are fused into $g_\eta = (L_\eta, C_\eta, A_\eta, \eta)$. $g_\eta$ is obtained by a product fusion if $\alpha$ is contained in probability parfactors, and by a summation fusion otherwise. A parfactor $g_{\eta'} = (L_{\eta'}, C_{\eta'}, A_{\eta'}, \eta')$ is then calculated by maximizing-out the entries of $\alpha$, as follows

$$\eta' = \left(\max_\alpha \eta\right)^{|L_\eta^\alpha : C|} \quad \text{in probability parfactors} \quad (11)$$

$$\eta' = \left(\max_\alpha \eta\right) \cdot |L_\eta^\alpha : C| \quad \text{in utility parfactors} \quad (12)$$

---
[1] If this is not the case, a "stub" parfactor $\eta(\alpha)$ is added to the model, such that $\alpha$ will then be contained in both types of parfactors. All table entries in a stub probability parfactor are 1, and all table entries in a stub utility parfactor are 0.

**Table 2:** *Eliminating rv formula $q(X,Y)$ by inversion*

| $p(X)$ | $q(X,Y)$ | $\phi$ |
|---|---|---|
| 0 | 0 | 0.2 |
| 0 | 1 | 0.3 |
| 1 | 0 | 0.7 |
| 1 | 1 | 0.7 |

| $r(Y)$ | $q(X,Y)$ | $\mu$ |
|---|---|---|
| 0 | 0 | 5 |
| 0 | 1 | 7 |
| 1 | 0 | 1 |
| 1 | 1 | 3 |

| $p(X)$ | $\phi'$ |
|---|---|
| 0 | $0.5^{|Y|}$ |
| 1 | $1.4^{|Y|}$ |

| $p(X)$ | $r(Y)$ | $\mu'$ |
|---|---|---|
| 0 | 0 | 6.2 |
| 0 | 1 | 2.2 |
| 1 | 0 | 6 |
| 1 | 1 | 2 |

**Table 3:** *Eliminating decision formula $d(X,Y)$ by inversion*

| $e(X)$ | $d(X,Y)$ | $\phi$ |
|---|---|---|
| 0 | 0 | 0.3 |
| 0 | 1 | 0.6 |
| 1 | 0 | 0.8 |
| 1 | 1 | 0.1 |

| $e(X)$ | $d^{max}(X,Y)$ | $\phi'$ |
|---|---|---|
| 0 | 1 | $0.6^{|Y|}$ |
| 1 | 0 | $0.8^{|Y|}$ |

$L_\eta^\alpha$ depicts the set of logical variables which are unique to $\alpha$ in $g_\eta$, $A_{\eta'} = A_\eta \setminus \{\alpha\}$, $L_{\eta'} = L_\eta \setminus L_\eta^\alpha$, and $C_{\eta'} = C_\eta{\downarrow}_{L_{\eta'}}$. The assignment to $\alpha$ which formed each entry in $\eta'$ is recorded for a backward phase. Finally, a residual model $G'$ is obtained by replacing $g_\eta$ with $g_{\eta'}$.

Let us examine model $\phi(e(X), d(X,Y))$, where both $e$ and $d$ are decision atoms. $d(X,Y)$ is eliminated from the model by calculating $\phi' = \max_d \phi^{|Y|}$, and recording the assignments to $d$ which yield the result entries. The exponentiation in $|Y|$ is the result of logical variable $Y$ being removed from the parfactor. The example is illustrated in Table 3.

## 6 Experimental Evaluation

We present results of three sets of experiments, all conducted on a E7400 Intel duo core machine, with 2.8GHz CPU speed and 3Gb of RAM. The propositional variable elimination for MEU was implemented in Java, with emphasis on performance, using a minimum deficiency heuristics [1] for variable ordering. Our lifted inference implementation is based on the Bayesian Logic Inference (BLOG) Engine, as found in http://people.csail.mit.edu/milch/blog/index.html, and was implemented in Java as well. Joint formula conversions were injected manually, prior to running the inference task.

Figure 3 depicts the results of lifted probabilistic inference in model $\phi(p(X), q(X), r(Y), s(Y))$. As can be seen, without joint formulas the model resorts to propositional inference and the problem becomes intractable. By introducing the joint formula $j(X) = \langle p(X), q(X) \rangle$, the problem is quickly solved. Figure 4 compares the results of propositional MEU vs. lifted MEU, in model $\phi_1(p(Y), q(X,Y), d(Z))$, $\phi_2(e(X), r(X)), \mu(e(X), q(X,Y))$, where $d$ and $e$ are decision atoms. Here, as in other FOVE variants, computation time is polynomial in the varying sizes of the domain, whereas computation time for the propositional algorithm is exponential in the size of the domain.

In Figure 5, three inference methods are compared: propositional inference, lifted inference, and lifted inference with joint formulas. Here, the propositional algorithm outperforms the lifted algorithm, but with the addition of joint formulas, the lifted algorithm outperforms the propositional algorithm, similarly to previous figures. A closer examination reveals the reason. The input model contains parfactors $\phi(d(X), e(X), p(X))$, $\mu_1(q(X,Y_1), q(X,Y_2), p(X))$ and $\mu_2(e(X), r(X), f(X))$, where $d$, $e$ and $f$ are decision atoms. Elimination of $r(X)$ by inversion, followed by the elimination of $f(X)$ by inversion, results in parfactor $\mu'_2(e(X))$. Two counting conversions of $q$ instances over $Y_1$ and $Y_2$, result in parfactor $\mu'_1(\#_{Y_1}[q(X,Y_1)], p(X))$, where $\#_{Y_1}[q(X,Y_1)]$ is then eliminated by inversion to construct parfactor $\mu''_1(p(X))$. Since $p(X)$ is included in both $\phi$ and $\mu''_1$, its elimination by inversion converts both parfactors into $\phi'(d(X), e(X))$, and $\mu'''_1(d(X), e(X))$.

At this phase, the decision atoms $d(X)$ and $e(X)$ cannot be eliminated by inversion, since they both reside in probability parfactors as well as in utility parfactors. Moreover, the fact that $d(X)$ appears with $e(X)$ in the same parfactor, prevents a counting conversion of both $d(X)$ and $e(X)$. The lifted algorithm resolves this conflict by grounding all the instances of the decision atoms, and continuing with a propositional model. However, the propositional algorithm was implemented much more efficiently than the lifted algorithm, which accounts for the performance gap between the two implementations. Once a joint formula $j(X) = \langle d(X), e(X) \rangle$ replaces all instances of $d$ and $e$, the $X$ logical variable could be counted out, resulting in instances of $\#_X[j(X)]$, and in an efficient lifted inference.

## 7 Conclusion

We introduced a novel contribution to the field of lifted inference, a model conversion method called joint formula conversion, and a following contribution which extends the counting conversion procedure. We then demonstrated how the new methods accelerates the task of lifted inference in some models. The use of joint formulas need not be limited to exact inference. In fact, we believe that the notion

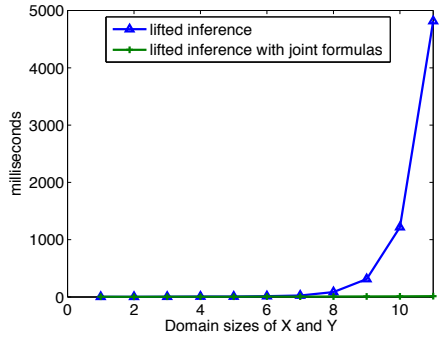 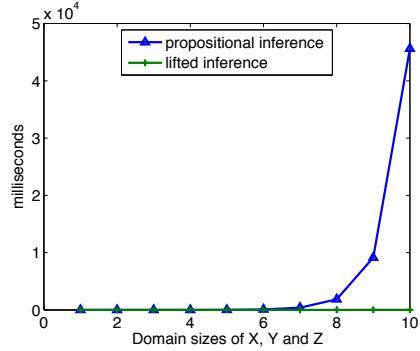 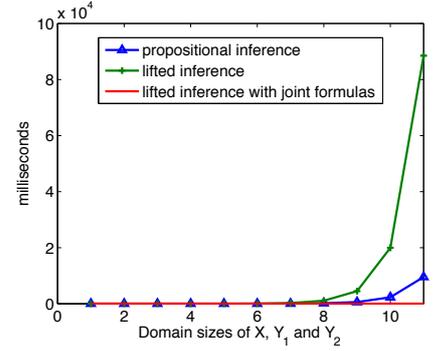

**Figure 3:** *Inference in model* $\phi(p(X), q(X), r(Y), s(Y))$

**Figure 4:** *MEU in model* $\phi_1(p(Y), q(X,Y), d^*(Z))$ $\phi_2(e^*(X), r(X))$ $\mu(e^*(X), q(X,Y))$ *(decision atoms in asterisk)*

**Figure 5:** *MEU in model* $\phi(d^*(X), e^*(X), p(X))$, $\mu_1(q(X,Y_1), q(X,Y_2), p(X))$ $\mu_2(e^*(X), r(X), f^*(X))$ *(decision atoms in asterisk)*

of joint formulas is generic enough to be adopted by some other relational models, such as relational MDP.

Our second contribution, the C-FOVE adaptation for MEU, is the first algorithm, to the best of our knowledge, to lift MEU computation. One interesting aspect of lifted MEU is that it generalizes many common probabilistic inference tasks. MPE and belief assessment, for instance, are both private cases of MEU computation, but more importantly – lifted MAP estimation, which has yet to be introduced, can be defined as a private case of lifted MEU, where the computational model contains only probability parfactors.

**Acknowledgements**

We thank the anonymous reviewers for their comments and useful suggestions. The authors were partly supported by ISF Grant 1101/07, the Paul Ivanier Center for Robotics Research and Production Management, and the Lynn and William Frankel Center for Computer Science.